\documentclass[conference,letters]{IEEEtran}
\IEEEoverridecommandlockouts
\usepackage{amsmath}
\usepackage{mathtools}
\usepackage{color}
\usepackage{array}
\usepackage{stfloats}
\usepackage{amsthm}
\usepackage{amssymb}
\usepackage{float}
\usepackage{nccmath}
\usepackage{graphicx}
\usepackage[linesnumbered,ruled]{algorithm2e}
\usepackage{setspace}
\usepackage{booktabs}
\usepackage{subcaption}
\usepackage{mwe}
\usepackage{cite}
\usepackage{amsmath,amssymb,amsfonts}
\usepackage{graphicx}
\usepackage{textcomp}
\usepackage{xcolor}

\usepackage{algorithmicx}
\usepackage{xspace}
\usepackage{graphicx}  
\usepackage{subcaption}
\newcommand{\argmin}{\mathop{\mathrm{argmin}}\limits} 
\newcommand{\argmax}{\mathop{\mathrm{argmax}}\limits} 
\newcommand{\BfPara}[1]{{\noindent\bf#1.}\xspace}
\def\BibTeX{{\rm B\kern-.05em{\sc i\kern-.025em b}\kern-.08em
    T\kern-.1667em\lower.7ex\hbox{E}\kern-.125emX}}
\begin{document}

\title{Adversarial Imitation Learning via Random Search}
%\\ \thanks{This work is jointly supported by Institute for Information \& Communications Technology Promotion (IITP) grant funded by the Korea government (MSIT) (No.2018-0-00170, Virtual Presence in Moving Objects through 5G); and National Research Foundation of Korea (2017R1A4A1015675).}
%}

\author{\IEEEauthorblockN{MyungJae Shin}
\IEEEauthorblockA{\textit{School of Computer Science and Engineering} \\
\textit{Chung-Ang University}\\
Seoul, Korea \\
mjshin.cau@gmail.com}
\and
\IEEEauthorblockN{Joongheon Kim}
\IEEEauthorblockA{\textit{School of Computer Science and Engineering} \\
\textit{Chung-Ang University}\\
Seoul, Korea \\
joongheon@gmail.com}
}

\maketitle 

\begin{abstract}
Developing agents that can perform challenging complex tasks is the goal of reinforcement learning. The model-free reinforcement learning has been considered as a feasible solution. However, the state of the art research has been to develop increasingly complicated techniques. This increasing complexity makes the reconstruction difficult. Furthermore, the problem of reward dependency is still exists. As a result, research on imitation learning, which learns policy from a demonstration of experts, has begun to attract attention. Imitation learning directly learns policy based on data on the behavior of the experts without the explicit reward signal provided by the environment. However, imitation learning tries to optimize policies based on deep reinforcement learning such as trust region policy optimization. As a result, deep reinforcement learning based imitation learning also poses a crisis of reproducibility. The issue of complex model-free model has received considerable critical attention. A derivative-free optimization based reinforcement learning and the simplification on policies obtain competitive performance on the dynamic complex tasks. The simplified policies and derivative free methods make algorithm be simple. The reconfiguration of research demo becomes easy. In this paper, we propose an imitation learning method that takes advantage of the derivative-free optimization with simple linear policies. The proposed method performs simple random search in the parameter space of policies and shows computational efficiency. Experiments in this paper show that the proposed model, without a direct reward signal from the environment, obtains competitive performance on the MuJoCo locomotion tasks.

% Deep reinforcement learning에 대한 연구가 점점 발전함과 함께, complex tasks를 해결할 수 있는 정책을 trial and error를 통해서 학습하는 방법에 대한 연구가 이루어지고 있습니다. 그러나, 정책의 학습을 위한 좋은 reward function을 구성하는 것은 매우 어려운 일입니다; 그리고 그것은 학습 과정에서 정책이 local optima에 떨어지는 (fall into) 일과 같은 문제점들을 발생시킵니다. 그러므로, 정책을 reward signal에 접근(access to)하는 것 없이 전문가의 demonstration으로 부터 정책을 학습하는 imitation learning에 대한 연구가 주목을 받기 시작했습니다. Imitation learning은 환경이 제공하는 명시적인 reward signal 없이 전문가의 행동에 대한 데이터를 바탕으로 직접적으로 정책을 학습합니다. 그러나, 최근 deep neural network 기반의 common imitation learning 은 optimization problems 를 가진다. which are highly non-convex, leading many methods to find suboptimal local optima; and it is  not robust to changes in hyperparameters, random seeds, or even different implementations of the same algorithm. 그 결과 학습 과정은 느리고 복잡해 집니다.
% 최근, 이러한 문제들을 해결하기 위해서, simplified reinforcement learning에 대한 연구로 simple linear policies based derivative-free optimization algorithm for training가 등장했고 shows a nearly optimal policy for a challenging instance. 그리고, 그것은 at least 15 times more efficient than the fastest competing deep reinforcement learning methods.
% 이 논문에서 우리는 이 derivative-free 기반 모델의 장점을 취하며 전문가의 행동 데이터를 기반으로 complex tasks에서 뛰어난 performance를 보여주는 정책을 학습하는 imitation learning method를 제안합니다. 제안된 방법은 simple linear policy 기반의 derivative-free imitation learning 으로, 학습 과정에서 random search  in  the  parameter  space  of  policies 를 수행하며 computational efficiency를 보여줍니다. 
% 이 논문의 실험은 MuJoCo locomotion task에 대해서 환경의 직접적인 reward signal 없이 제안된 모델이 전문가의 정책을 기반으로 뛰어난 performance의 정책을 학습한 것을 보여줍니다. 

\end{abstract}

%\begin{IEEEkeywords}
%Reinforcement learning, generative adversarial network, imitation learning, random search, linear policy
%\end{IEEEkeywords}

\section{Introduction}\label{sec:1}
% \IEEEPARstart{A}{ccording}

%2013년, Deep Q-Learning 논문은 고전적인 Q-Learning과 신경 네트워크를 결합하여 Atari 게임을 성공적으로 해결하는 방법을 보여 주었으며 RL이 가장 주목받는 연구 분야로 다시 활력을 불어 넣었습니다~\cite{mnih2013playing}.
In 2013, the Deep Q-Learning showed how to combine classical Q-Learning with convolution neural network to successfully solve Atari games, reinvigorating reinforcement learning (RL) as one of the most remarkable research fields~\cite{mnih2013playing, levine2013guided, schulman2015trust, lillicrap2015continuous, horgan2018distributed}.
% 신경 네트워크를 통해 가치 기능을 근사하는 DRL (deep reinforcement learning)에 많은 관심이 모아졌습니다.
As a result, much attention has been drawn to deep reinforcement learning (DRL) that approximates value function through neural network.
% DRL 은 시스템 다이내믹 모델을 필요로하지 않고 dynamic taks를 제어하기위한 솔루션으로 여겨지고 있습니다.
DRL has been studied as a feasible solution for controlling dynamic tasks (i.e., autonomous driving, humanoid robot, etc) without requiring models of the system dynamics~\cite{busoniu2008comprehensive, lange2012autonomous, kober2012reinforcement, zhu2018dexterous,chen2015deepdriving}. 
% ===================== 일반적인 DRL 의 문제점 제시 (1) =================================
% 모델의 복잡성 => 재현 불가능
% 환경에 민감 

% 그러나, 연구자들이 직면하게 된 주요한 문제는 DRL이 reasonable performance를 위해서 너무 많은 데이터가 필요하다는 것입니다. 
However, the main challenge faced by many researchers is the model-free RL requires too much data to achieve reasonable performance~\cite{schulman2015trust, schulman2017proximal}. 
% 이를 해결하기 위해서, 모델이 복잡 해지고 재현성 위기가 발생했습니다.
To address this issue, the models become complicated; and the models lead to reproducibility crisis. 
% 더 나아가, Model들이 같은 알고리즘의 다른 구현과 hyperparameter에 민감합니다. 
Furthermore, the models are sensitive to the implementation structure of the same algorithm and rewards from environments.
% 그 결과 reconstruction result들이 연구 데모와 달리 reasonable performance를 보여주지 못하거나 sub-optimal에 빠졌습니다. 
As a result, the reconstruction results do not show reasonable performance, and stuck in sub-optimal.
% 그래서, 아직 실제로 적용이 되지 못하고 있습니다. 
Therefore, the models have not yet been successfully deployed to control systems~\cite{henderson2017deep, islam2017reproducibility}.

% ================== 환경에 민감한 문제점 보충 설명 (2) ====================================
% 
The rewards sensitivity makes the optimization of model-free RL to be difficult.
% 보상 신호 (환경 응답)는 네트워크를보다 잘 제어하기 위해 개선하는 방법에 관한 정보입니다.
The reward signals are information about how to improve the inner neural network to better control. 
%문제는 내부 신경 네트워크의 최적화는 네트워크 가중치의 영향을 최적화에 전파할지 여부를 결정할 때 환경의 reward signals 결과에 의존한다는 것과 관련이 있습니다.
The optimization of the inner network depends on the reward signals in determining whether to propagate the effects of network weights to optimization.  
% 많은 dynamic control tasks에서, reward signals는 극단적으로 희박하거나 전혀 없습니다.
In many dynamic tasks, the reward signals are extremely sparse or none at all. 
% 그 결과, 모델들이 sub-obtimal에 빠지거나 합리적인 성능에 도달하지 못하는 경우에 효과적인 해결 방법이 없었습니다.
As a result, when the models are stuck in sub-optimal, the problems can not be handled appropriately.

% ====================== sub-obtimal에 빠지는 것과 같은 문제 해결 방법 (1) =======================
% Imitation Learning 
% BC, GAIL 
% GAIL은 DRL 기반 
% 여전히 재현성의 위기가 존재
% DRL의 재현성 위기를 다루기 위한 연구 ES와 ARS
% ARS
% 이 논문에서의 제시 연구 언급

% 동적 시스템에서 합리적인 성능을 이끌어 내기 위해 신호 밀도를 높여주는 보상 형성이 연구되었습니다. 
The reward shaping which makes the signals to be more dense to lead to the reasonable performance in dynamic systems has been studied. Several attempts have been made to manually design reward function by hand. However, it is difficult to configure an appropriate reward function by hand. Therefore, imitation learning is proposed. Imitation learning trains the models based on the desired behavior demonstrations rather than configuring the reward function that would generate such behavior.
% 그 결과, Imitation learning은 sparse reward와 같은 문제들로부터 생기는 sub-optimal과 같은 문제가 있을 때 useful 했습니다. 
Imitation learning shows impressive performance when there is sub-optimal problems arising from problems such as sparse reward. 
% Imitation learning은 로봇 공학과 자율 주행과 같은 분야에서 주목할 만한 성과를 보였습니다~\cite{billard2008robot, pomerleau1991rapidly, pomerleau1989alvinn}.
Imitation learning has performed remarkably well in areas such as robotics and autonomous navigation~\cite{billard2008robot, pomerleau1991rapidly, pomerleau1989alvinn}.
% Imitation learning에서, 전문가의 demonstartion을 통한 감독은 지침이됩니다.
In imitation learning, supervision through a set of expert demonstrations is to be a guideline which learner can query when the models are trained.
The simplest method of imitation learning is behavioral cloning (BC). It works by collecting training data from the expert demonstrations, and then uses it to directly learn a policy. 
% 데이터가 풍부 할 때 효율성은 높지만 에이전트가 전문적인 궤도(trajectories)에서 벗어나면 에이전트가 연약한 경향이 있습니다.
BC shows high performance when we have abundant expert demonstrations, but agents tend to be fragile if the agents deviate from trajectories which trained in training procedure. 
% 이것은 supervised learning 이 agent의 trajectories 전체가 아니라 학습 데이터의 1-step deviation error를 줄이는 것을 목표로 하기 때문입니다. 
This is because supervised learning method tries to reduce the 1-step deviation error of training data, not to reduce the error of entire trajectories.
% =================================================================================================
% 최근 에이전트의 distribution of state-action trajectories가 전문가의 distribution of expert trajectories와 일치 하도록 에이전트를 학습하는 방법으로 model-free imitation learning called GAIL 이 제안되었습니다. 
Recently, as the method that makes the distribution of state-action trajectories of the agents to be matched the distribution of expert trajectories of the experts, a model-free imitation learning called GAIL (Generative Adversarial Imitation Learning) is proposed~\cite{ho2016generative}.
% Generative Adversarial Networks (GAN)를 활용하는 GAIL은 cost function을 학습하는 대신 전문가의 demonstartion의 행동과 learner의 행동을 구분하여 보상을 제공하는 adversarial network를 학습하고 이를 기반으로 시뮬레이트 된 루프에서 롤아웃을 수행하는 내부 루프의 RL 절차를 수행함으로써 supervised learning(i.e., behavior cloning)의 특정 궤도에 집중되는 한계를 완화합니다. 
% GAIL은 학습 과정에서 reasonable performance를 얻기 위해서 환경과의 상호작용이 많이 필요합니다.
% 그리고 그 과정에서 안정적인 학습을 위해서 TRPO, PPO 와 같은 복잡한 DRL 알고리즘이 사용됩니다.
% 그로인해, GAIL 역시 재현가능성의 위기를 이끕니다. 
In GAIL, the discriminator of Generative Adversarial Networks (GAN) takes a role of the reward function. The reward signals from the discriminator means the probability that how much the learner's trajectories is similar to the trajectories of expert. By using this reward, GAIL train the policy of agent based on trust region policy optimization (TRPO). Through GAIL, the agent learns the policy which achieves the expected rewards of expert trajectories in dynamic continuous tasks and sometimes better than the experts, because deep reinforcement learning is executed in inner loop and it’s not constrained to always be close to the expert~\cite{ho2016generative}. GAIL requires a lot of interaction with the environment to get reasonable performance in the training procedure. Therefore, to stabilize the training, the complex DRL algorithms (i.e., TRPO and proximal policy optimization ) are used. As a result, GAIL also poses a crisis of reproducibility.
% ================= 재현 가능성을 해결할 수 있는 연구 ===================================
% ARS 
To address the crisis, the simplest model-free RL has been studied. Recently, diﬀerent directions have been proposed. Firstly, evolution strategies (ES) shows a powerful derivative-free policy optimization method~\cite{es}. Secondly, the simple linear policy based on the natural gradient policy algorithm shows the competitive performance on the MuJoCo locomotion tasks~\cite{rajeswaran2017towards}. 
Augmented random search (ARS) is proposed as a result of integrating these concepts. In ARS, the policy is trained through random searches in the parameter space. ARS is a derivative-free simple linear policy optimization method~\cite{ars}. 
% ===================== 이 연구에서 제시하는 연구 방법 제시=========================================== 
The specific objective of this study is to propose highly reproducible imitation learning method.
In this work, we combine ideas from the work of \cite{ars} and \cite{ho2016generative}. The simple linear policies are used. By using the derivative-free random search, the trained policies show stabilized reasonable performance. Furthermore, the discriminator is used to replace the reward function of environments. The trained agent achieves the expected rewards of expert trajectories. Through the experiments, we demonstrate that a simple random search based imitation learning method can train linear policy efficiently on MuJoCo locomotion benchmarks. 
For more details, our contributions are as follows:
\begin{enumerate}
    \item The performance of our method on the benchmark MuJoCo locomotion tasks. Our method can successfully imitate expert demonstration; and static and linear policies can achieve high rewards on all MuJoCo task.
    \item Since previous imitation learning methods is based on RL methods which has complicate configuration to handle the complex tasks, it difficult to choose what is the best method for a specific task as well as to reconstruct the result. Howver, our method is based on the derivate-free simple random search algorithm with simple linear policies; and thus it can solve a reproducibility crisis. 
    \item Within the knowledge we know, the combination of adversarial network and simple random search is the state-of-the-art; thus the proposed method will be a new guideline of imitation learning in the future. 
\end{enumerate}

% Section 설명 ==========================
Sec.~\ref{sec2} describes the background knowledge and Sec.~\ref{sec3} shows how to design the proposed algorithm. Sec.~\ref{sec4} shows the experiment results of the proposed imitation learning on expert demonstration in the MuJoCo locomotion environment.
Sec.~\ref{sec:final} concludes this paper and presents future work.

% Section 2에서는 이 논문에서 제안하는 알고리즘을 이해하기 위한 배경 지식에 대한 설명을 합니다. Section 3에서는 제안되는 알고리즘의 설계 및 구현 방법에 대해서 보여주며 Section 4에서는 Mujoco locomotion 환경에서 전문가 데이터를 기반으로 학습한 결과를 보여줍니다. Section 5에서는 제안된 알고리즘에 대한 discusstion와 논문의 결론을 이야기합니다. 

\section{Background}\label{sec2}
\BfPara{Preliminaries} A Markov Decision Process (MDP) is defined as $M = \left\{\mathcal{S}, \mathcal{A}, T, r\right\}$, where $\mathcal{S}$ denotes the state space, $\mathcal{A}$ denotes the set of possible actions, $T$ denotes the transition model and $r$ denotes the reward structure. Throughout the paper, we consider a finite state space $\mathcal{S} \in \mathbb{R}^n$ and a finite action space $\mathcal{A} \in \mathbb{R}^p$. The goal of imitation learning is to train a policy $\pi_\theta  \in \Pi : \mathcal{S} \times \mathcal{A} \to \left[ 0, 1 \right]^p$ which can imitate expert demonstration using the idea from generative adversarial network (GAN) $\mathcal{D}_\phi(s, a) \to \left[ 0, 1 \right]$ where $\theta \in \mathbb{R}^n$ are the policy parameters and $\phi \in \mathbb{R}^{n+p}$ are the discriminator parameters~\cite{ho2016generative}.

Expert demonstrations  $\mathcal{T}_E = \left\{\tau_1, \tau_2, ..., \tau_N \right\}$ is available. Each demonstration $\tau_i$ is consist of a set of action state-action pairs such that $\tau_i = \left\{ (s_0, a_0), (s_1, a_1), \dots , (s_T, a_T) \right\}$ where $N$ is the number of demonstration set and $T$ is the length of episode.

\BfPara{Behavior Cloning (BC)}
Behavioral cloning learns a policy as a way of supervised learning over state-action pairs from expert demonstration. Distribution of states which is visited by expert is defined as $P_E = P(s|\pi_E)$. The objective of BC is defined as: 
\begin{equation}
    \label{bc_objective}
    \argmin_{\theta} \mathbb{E}_{s\thicksim P_E}\left[ \mathcal{L}\left( a_E, \pi_\theta\left(s\right)\right)\right] = \mathbb{E}_{s\thicksim P_E}\left[ \left( a_E - \pi_\theta\left(s\right)\right)^2\right]
\end{equation}
Though BC is appealingly simple  (\ref{bc_objective}), it only tends to trains polices successfully when we have large amounts of expert data. BC tries to minimize 1-step deviation error along the expert demonstration; it makes the trained polices to be fragile when distribution mismatch between training and testing. In the some case of experiments, by initializing policy parameters with BC, the learning speed of the proposed method is improved; and thus BC is adapted to our evaluation. 

\BfPara{Inverse Reinforcement Learning (IRL)}  Inverse reinforcement learning is able to learns a policy in the case that a MDP specification is known but the reward $r$ is unknown and expert demonstrations $\mathcal{T}_E$ is available. IRL uncovers the hidden reward function $R^*$ which explains the expert demonstration. 
\begin{equation}
    \label{irl_reward}
    \mathbb{E}\left[\sum_{t=0}^{\infty}\gamma^tR^*(s_t) | \pi_E \right] \ge \mathbb{E}\left[\sum_{t=0}^{\infty}\gamma^tR^*(s_t) | \pi_\theta \right] 
\end{equation}
Based on the uncovered reward function $R^*$, the reinforcement learning is carried out to train the policy $\pi_{\theta}$. The objective of IRL can be defined as:
\begin{equation}
    \label{irl_rl}
    \argmax_{\theta} \mathbb{E}_{s\thicksim P_E}\left[ R^*(s, \pi_\theta(s))\right] 
\end{equation}

IRL learns a reward function $R^*$ that explains entire expert trajectories. Therefore, a problem which makes the trained policy to be fragile when there are mismatch between training and testing environment is not an issue. However, IRL is expensive to run because it has to perform both reward function optimization (\ref{irl_reward}) and policy optimization (\ref{irl_rl}) at the same time. 

\BfPara{Generative Adversarial Imitation Learning (GAIL)~\cite{ho2016generative}}
Generative adversarial imitation learning (GAIL) learns a policy that can imitate expert demonstration using the adversarial network from generative adversarial network (GAN). The objective of GAIL is defined as: 
\begin{equation}
        \label{gail_loss}
        \argmin_\theta \argmax_\phi \mathbb{E}_{\pi_\theta} \left[ \log \mathcal{D}_\phi(s, a)\right]+\mathbb{E}_{\pi_E} \left[ \log(1-\mathcal{D}_\phi(s,a))\right]
\end{equation}
where $\pi_\theta, \pi_E$ are a policy which is parameterized by $\theta$ and an expert policy. $\mathcal{D}_\phi(s, a) \to \left[ 0, 1 \right]$ is an discriminator parameteriezd by $\phi \in \mathbb{R}^{n+p}$~\cite{ho2016generative}. The discriminator network and policy play an min-max game to train policy $\pi_\theta$ by having the policy $\pi_\theta$ confuse a discriminator $\mathcal{D}_\phi$. Discriminator $\mathcal{D}_\phi$ uses state-action pair $\tau_i$ from the expert demonstrations $\mathcal{T}_E$; it distinguish between the expert trajectories and the trajectories distribution of the trained policy. $\mathcal{D}_\phi(s, a)$ is the probability that state-action pairs $(s,a)$ belongs to an expert demonstration. During the policy optimization, the GAIL uses trust region policy optimization (TRPO) to prevent perturbations of policy. Let the objective loss Equation (\ref{gail_loss}) as $\mathcal{L}_{G}$. The gradient of each component is as follows:
\begin{align}
\begin{split}
    \label{gail_diff_phi}
    \bigtriangledown_\phi \mathcal{L}_{G} ={}& \mathbb{E}_{\pi_\theta} \left[  \bigtriangledown_\phi\log \mathcal{D}_\phi(s, a)\right] \small{+} \mathbb{E}_{\pi_E} \left[  \bigtriangledown_\phi\log(1-\mathcal{D}_\phi(s,a))\right]
\end{split}\\
\begin{split}\label{gail_diff_theta}
    \bigtriangledown_\theta \mathcal{L}_{G} ={}& \mathbb{E}_{\pi_\theta} \left[  \bigtriangledown_\theta\log \mathcal{D}_\phi(s, a)\right]
\end{split}\\
\begin{split}\label{gail_scorefunction}
      ={}& \mathbb{E}_{\pi_\theta} \left[ \bigtriangledown_\theta\log \pi_\phi(a|s)Q(s,a)\right]
\end{split}
\end{align}

In Equation (\ref{gail_diff_theta}), $\log \mathcal{D}_\phi (s, a)$ can not be differentiable with respect to $\theta$. Therefore, the form of the policy gradient Equation (\ref{gail_scorefunction}) is used to compute the gradient. The discriminator $\mathcal{D}_\phi$ takes the role of a reward function; and thus it gives learning signal to the policy~\cite{ho2016generative, gasil, optiongan}.

\SetInd{0.5em}{0.8em}
\begin{algorithm}[t!]
\setstretch{1.2}
        \SetKwInOut{Hyperparameters}{Hyperparameters}
        \SetKwInOut{Initialize}{Initialize}
        \Hyperparameters{$\alpha$  step size, $N$  number of sampled directions per iteration, $\delta$  a zero mean Gaussian vector, $\nu$ a positive real number standard deviation of the exploration noise}
        \Initialize{$\theta_0 = \bold{0} \in \mathbb{R}^{p \times n}, \mu_0 = \bold{0}\in \mathbb{R}^n,$ and $\sum_{0} = \bold{I}_n \in \mathbb{R}^{n\times n}$}
        \While{$t \le$ Max Iteration}
        {
        Sample $\bold{\delta_t} = \left\{ \delta_1, \delta_2, . . . , \delta_N ; \delta_i \in \mathbb{R}^{p \times n} \right\} $ with i.i.d.\\
        Collect $2N$ rollouts and their corresponding rewards using the $2N$ policies.\\
        \Indp 
        {
        $\pi_{t,i,+(x)} = (\theta_t + \nu\delta_i) diag (\sum_t)^{−1/2}(x - \mu_t )$\\
        $\pi_{t,i,-(x)} = (\theta_t - \nu\delta_i) diag (\sum_t)^{−1/2}(x - \mu_t )$\\
        for $i \in  \left\{ 1, 2, \dots, N \right\}$\\
        }
        \Indm
        Update Step:
        $\theta_{t+1} = \theta_{t} + \frac{\alpha}{N\sigma_R}\sum^{N}_{i=1}{\left[ r(\pi_{t,(i),+}) - r(\pi_{t,(i),-}) \right] }$\\
        Set $\mu_{t+1}$, $\sum_{t+1}$ to be the mean and covariance of the states encountered from the start of training.\\
        $t$ = $t + 1$
        }
    \caption{Augmented Random Search $V2$}
    \label{algo:ARS_V2}
\end{algorithm}

\BfPara{Augmented Random Search (ARS)~\cite{ars}}
Augmented random search (ARS) is a model-free reinforcement learning algorithm based on random search in the parameter space of policies~\cite{matyas1965random, ars}. The objective of ARS is to learn the policy which maximize the expected rewards; it can be described:
\begin{equation}
    \label{brs_objective}
    \max_{\theta \in \mathbb{R}^n} \mathbb{E} \left[ r(\pi_\theta)\right]
\end{equation}
where $\theta$ is parameter of the linear policy $\pi_\theta : \mathbb{R}^n \to \mathbb{R}^p$.

The random search in parameter space makes the algorithm to be derivative-free optimization with noise~\cite{matyas1965random, ars}. Random search algorithm which is the basic concept of ARS selects directions uniformly in parameter space and updates the policies along the selected direction without using a line search. 
For updating the parameterized policy $\pi_\theta$, the update direction is calculated as follow: 
\begin{equation}
    \label{brs_dircetion}
    \frac{r(\pi_{\theta - \nu\delta}) + r(\pi_{\theta + \nu\delta})}{\nu},
\end{equation}
for $\delta$ a zero mean Gaussian vector and $\nu$ a positive real number standard deviation of the exploration noise. When $\nu$ is small enough, $\mathbb{E}_\delta  \left[r(\pi_{\theta+\nu\delta})\right]$ can be the smoothed form of Equation (\ref{brs_objective}). Therefore, an update increment is an unbiased gradient estimator with respect to $\theta$ of $\mathbb{E}_\delta  \left[r(\pi_{\theta+\nu\delta})\right]$; and it makes the update step of the policies $\pi_\theta$ to be unbiased update~\cite{ars, nesterov2011random}. Based on this fact, Bandit Gradient Descent which is called BRS was proposed in~\cite{flaxman2005online}. Let the $\theta_t$ is the weight of policy at $t$-th training iteration. $N$ denotes that the number of sampled directions per iteration. In BRS, the update step is configured as follows:
\begin{equation}
    \label{brs}
    \theta_{t+1} = \theta_{t} + \frac{\alpha}{N} \sum_{i=1}^{N}{\left[r(\pi_{\theta + \nu\delta_i}) - r(\pi_{\theta - \nu\delta_i})\right]\delta_i}
\end{equation}

However, the problem of random search in the parameter space of policies is large variations in terms of the rewards $r(\pi_\theta \pm \nu\delta)$ which are observed during training procedure. The variations makes the updated policies to be perturbed through the updates step (\ref{brs}). To address the large variation issue, the standard deviation $\sigma_R$ of the rewards which are collected at each iteration is used to adjust the size of the update step in ARS. Based on the adaptive step size, the ARS shows higher performance compared to the deep reinforcement learning algorithms (i.e., PPO, TRPO, A3C, etc.) and BRS even if the simple linear policy is used. In this paper, for policy optimization of imitation learning, the ARS V2 algorithm is used as a baseline. The ARS algorithm is described as Algorithm~\ref{algo:ARS_V2}.

The update step of ARS means that if $r(\pi_{t,(i),+}) > r(\pi_{t,(i),-})$, the policy weights $\theta_t$ is updated in the direction of $\delta_i$. However, if $r(\pi_{t,(i),+}) < r(\pi_{t,(i),-})$, the policy weights $\theta_t$ is updated in the direction of $- \delta_i$. This update step does not need backpropagation procedure which is used to optimize DRL; and thus ARS is derivative-free optimization. Furthermore, ARS shows that simple linear policies can obtain competitive performance on the high dimensional complex problems, showing that complicated neural network policies are not needed to solve these problems~\cite{ars, rajeswaran2017towards}.

\section{Random Search based Imitation Learning}\label{sec3}
The proposed simple random search based adversarial imitation learning (AILSRS) is based on the ARS-V2 and generative adversarial imitation learning (GAIL)~\cite{ho2016generative,ars}. The main idea of AILSRS is to update the linear policy $\pi_\theta$ to imitate expert trajectories using adversarial network. We describe the details of AILSRS in Section \ref{sec3} and make a connection between adversarial network and ARS. 
\begin{figure}[t!]
    \centering
    \includegraphics[width=0.9\columnwidth]{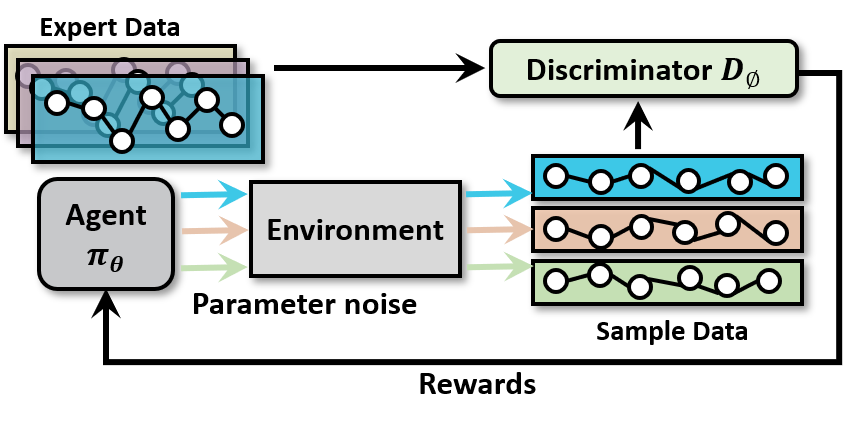}
    \caption{Structure of AILSRS.}
    \label{fig:structure}
\end{figure}

\BfPara{Updating discriminator ($\mathcal{D}_\phi)$} In GAIL, Equation (\ref{gail_loss}) draws a connection between adversarial network and imitation learning algorithm. The policy $\pi_\theta$ is trained to confuse a discriminator $\mathcal{D}_\phi$. The $\mathcal{D}_\phi$ tries to distinguish between the distribution of trajectories which is sampled by the policy $\pi_\theta$ and the expert trajectories $\mathcal{T}_E$. The trajectories are consist of state-action pair $(s, a)$. The discriminator takes the role of a reward function in AILSRS shown in Fig. \ref{fig:structure}; and thus the result of the discriminator is used to train the policy $\pi_\theta$. Therefore, the performance of the discriminator is important in our method. However, since the policy $\pi_\theta$ is updated every iteration, sampled trajectories which are used to train the discriminator are changed. The training of the discriminator is not stabilized; and thus it makes the inaccurate reward signal. As a result, the policy is perturbated during update step~\cite{sorg2010reward, gasil}. 
In AILSRS, the loss function of least square GAN (LS-GAN) is used to train a discriminator $\mathcal{D}_\phi$~\cite{mao2017least}. The objective function of the discriminator is as follows:
\begin{equation}
  \label{lsgan_loss}
  \begin{multlined}
     \argmin_\phi L_{LS}(\mathcal{D}) = \frac{1}{2}\mathbb{E}_{\pi_E} \left[(\mathcal{D}_\phi(s, a) - b)^2\right]+\\
     \frac{1}{2}\mathbb{E}_{\pi_\theta} \left[ (\mathcal{D}_\phi(s,a) - a)^2\right]
\end{multlined}
\end{equation}
where $a$ and $b$ are the target discriminator labels for the sampled trajectories from the policy $\pi_\theta$ and the expert trajectories. In Equation (\ref{gail_loss}), sampled trajectories which are far from the expert trajectories but on the correct side of the decision boundary are almost not penalized by sigmoid cross-entropy loss. In a contrast, the least-squares loss function (\ref{lsgan_loss}) penalizes the sampled trajectories which are far from the expert trajectories on either side of decision boundary~\cite{mao2017least}. Therefore, the stability of training is improved; and it leads the discriminator to give accurate reward signals to the update step. In LS-GAN, $a$ and $b$ have relationship $b-a=2$ for Equation (\ref{lsgan_loss}) to be Pearson $\mathcal{X}^2$ divergence~\cite{mao2017least}. However, we use $a=0$ and $b=1$ as the target discriminator labels. The result of the discriminator $\mathcal{D}_\phi$ in the range of 0 to 1. These values are chosen by empirical results.  

\SetInd{0.5em}{0.5em}
\begin{algorithm}[t!]
\setstretch{1.2}
        \SetKwInOut{Hyperparameters}{Hyperparameters}
        \SetKwInOut{Initialize}{Initialize}
        \Hyperparameters{$\alpha$  step size, $N$  number of sampled directions per iteration, $\delta$  a zero mean Gaussian vector, $\nu$ a positive real number standard deviation of the exploration noise}
        \Initialize{$\theta_0 = \bold{0} \in \mathbb{R}^{p \times n}, \mu_0 = \bold{0}\in \mathbb{R}^n,$ and $\sum_{0} = \bold{I}_n \in \mathbb{R}^{n\times n}$}
        \While{$t \le$ Max Iteration}
        {
        Sample $\bold{\delta_t} = \left\{ \delta_1, \delta_2, . . . , \delta_N ; \delta_i \in \mathbb{R}^{p \times n} \right\} $ with i.i.d.\\
        Collect $2N$ rollouts and their corresponding rewards using the $2N$ policies.\\
        \Indp 
        {
        $\pi_{t,i,+(s)} = (\theta_t + \nu\delta_i) diag (\sum_t)^{−1/2}(s - \mu_t )$\\
        $\pi_{t,i,-(s)} = (\theta_t - \nu\delta_i) diag (\sum_t)^{−1/2}(s - \mu_t )$\\
        for $i \in  \left\{ 1, 2, \dots, N \right\}$\\
        }
        \Indm
        Update discriminator parameter $\phi_t$ :\\
         \Indp
         $\nabla_{\phi_t}L_{LS} = \frac{1}{2}\mathbb{E}_{\pi_E} \left[(\nabla_{\phi_t}\mathcal{D}_{\phi_{t}}(s, a) - b)^2\right]$\\
         \Indm
         \Indp\Indp\Indp\Indp\Indp\Indp\Indp\Indp $+\frac{1}{2}\mathbb{E}_{\pi_\theta}\left[(\nabla_{\phi}\mathcal{D}_{\phi_{t}}(s,a) - a)^2\right]$\\
         \Indm\Indm\Indm\Indm\Indm\Indm\Indm\Indm
        Update the policy parameter $\theta_t$ :\\
        \Indp
        $\theta_{t+1} = \theta_{t} + \frac{\alpha}{N\sigma_R}\sum^{N}_{i=1}{\left[ r(\pi_{t,(i),+}) - r(\pi_{t,(i),-}) \right]\delta_{(i)} }$\\
        \Indm
        where trajectories $T$ sampled from $\pi_{(t,(i),\pm)}$ \\ 
        and $r(\pi_{t,(i),\pm}) = \mathbb{E}_{(s,a)\thicksim \pi_{t,(i),\pm}} \left[ -\log( 1 - \mathcal{D}_{\phi_{t}}(T)) \right]$\\
        Set $\mu_{t+1}$, $\sum_{t+1}$ to be the mean and covariance of the states encountered from the start of training.\\
        $t$ = $t + 1$
        }
    \caption{Adversarial Imitation Learning through Simple Random Search (AILSRS)}
    \label{algo:AILSRS}
\end{algorithm}

\BfPara{Updating policy ($\pi_\theta$)}
The discriminator in AILSRS is interpreted as a reward function for which the policy optimizes. The form of reward signal is as follows:
\begin{equation}
  \label{reward_signal}
  \begin{multlined}
     r_{\pi_\theta}(s, a) = - \log(1 - \mathcal{D}_\phi(s, a))
\end{multlined}
\end{equation}

This means that if the trajectories sampled from the policy $\pi_\theta$ is similar to expert trajectories, the policy $\pi_\theta$ gets higher reward $r_{\pi_\theta}(s, a)$. The policy $\pi_\theta$  is updated to maximize the discounted sum of rewards given by the discriminator rather than the reward from the environment as shown in Fig.~\ref{fig:structure}. The objective of AILSRS can be described:
\begin{equation}
  \label{ailsrs_objective}
  \begin{multlined}
    \argmax_{\theta} \mathbb{E}_{(s, a)\thicksim \pi_\theta}\left[ r(s, a)\right]
    =\mathbb{E}_{(s, a)\thicksim \pi_\theta}\left[ - \log(1 - \mathcal{D}_\phi(s, a))\right]
  \end{multlined}
\end{equation}

This Equation (\ref{ailsrs_objective}) is connection of adversarial imitation learning and simple random search.

\BfPara{Algorithm}
Foremetioned, AILSRS is based on ARS which is model-free reinforcement algorithm. Therefore, AILSRS uses simple linear policy and parameter space exploration for derivative-free policy optimization. The parameters of policy $\pi_\theta$ is denoted $\theta$ and hence $\theta$ is a $p \times n$ matrix. The noises $\delta$ of parameter space for exploration are also $p \times n$ matrix. The noises are sampled from a zero mean and $\nu$ standard deviation Gaussian distribution. AILSRS algorithm is shown in Algorithm~\ref{algo:AILSRS}. For each iteration, the noises $\bold{\delta}$ which mean search directions in parameter space of policy are chosen randomly (line [2]). Each of the selected $N$ noises make two policies in the current policy $\pi_\theta$. We collect $2N$ rollouts and rewards from $N$ noisy policies $\pi_{t,i,\pm} = \theta_t \pm \nu\delta_i$ (line [3-6]). The high dimensional complex problems have multiple state components with various ranges; and thus it makes the policies to result in large changes in the actions when the same sized changes is not equally influence state components. Therefore, the state normalization is used in AILSRS (line [4-5,14]); and it allows linear policies $\pi_{t,i,\pm}$ to have equal influence for the changes of state components when there are state components with various ranges.~\cite{ars, es, nagabandi2018neural}. The discriminator $\mathcal{D}_\phi$ gives the reward signal to update step. However, since the trajectories for the training of the discriminator can only be obtained from current policies $\pi_{\theta_t}$, a discriminator is trained whenever the policy parameter $\theta_t$ is updated. The discriminator $\mathcal{D}_\phi$ finds the parameter $\phi$ which minimize the objective function (\ref{lsgan_loss}) (line [7-9]). By using the reward signals from the discriminator, the policy weight is updated in the direction of $\delta$ or $-\delta$ based on the result of $r(\pi_{t,(i),+}) - r(\pi_{t,(i),-})$ (line [10-13]). The state normalization is based on the information of the states encountered during the training procedure; and thus $\mu$ and $\sum$ are updated (line [14]).

\section{Experiments}\label{sec4}
% 2 page 이상 
% 실험 환경 및 구현은 tensorflow 
% 실험 셋팅 hyper parameter table 
% behavior cloning 방법 
% 실험 셋팅 mujoco 언급 
% mujoco가 어려운 점 continuos task 
% 6개 graph 
% expert 점수 & AILSRS 학습 곡선
% 설명
% 6개 graph 
% 10 20 40 으로 성능 뽑았는지 그래프 (대충해 대충) 
% 설명

\BfPara{Implementation Setting}
In this paper, adversarial imitation learning through simple random search (AILSRS) is implemented with Python/TensorFlow~\cite{tensorflow}. Multi-GPU platform (equipped with the 2 NVIDIA Titan XP GPUs using 1405\,MHz main clock and 12\,GB memory) was used for training and evaluation the proposed method. The performance of AILSRS is evaluated on the MuJoCo locomotion tasks~\cite{mujoco,gym}. The OpenAI Gym provides benchmark reward functions for Gym environments; and it is used to evaluate the performance of AILSRS. Evaluation on three random seeds is widely adopted in the researches. Therefore, We wanted to show the performance of AILSRS in an equally competitive position~\cite{ars,es,ho2016generative,gasil}. The experiment is implemented with 1, 3 and 5 random seeds of the MuJoCo tasks. The hyperparameters were summarized in Table \ref{hyperparameters}. Results show that AILSRS achieves rewards from expert trajectories in various random seed evaluation. Each training curve was smoothed through a Gaussian filter for the average of the experimental results. The discriminator network for AILSRS is consist of two hidden layer of 100 units, with $tanh$ non-linearities in between layers. In the experiment, BC used the same structure policy as AILSRS.
% Sample number based result ===================================================================

\begin{table}[t!]
\footnotesize
\caption{Hyperparameters}
\label{hyperparameters}
\begin{center}
    \scalebox{1}{
	\begin{tabular}{l|c}
    \toprule
    Hyperparameters & Descriptions \\
    \midrule [1.0pt]
    $\alpha$ update step & 0.02\\ 
    $N$ number of direction & 320\\
    $\nu$ standard deviation of noise & 0.03\\
    Max iteration of each rollout & 1000\\
    Max Training iteration & 100000\\
    Discriminator learning rate & 0.00025\\
    Discriminator batch size & episode length of each rollout\\
    Discriminator training iteration & 3\\[0.2ex] 
    \bottomrule
	\end{tabular}
	}
\label{tab:hresult}
\end{center}
\vspace{-5mm}
\end{table}

\begin{figure*}[t!] 

\begin{subfigure}{0.5\textwidth}
\includegraphics[width=1\columnwidth]{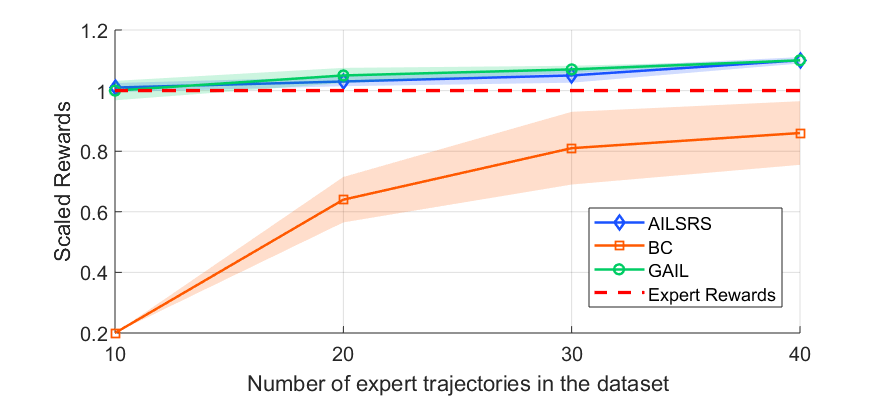}
\caption{HalfCheetah-v2} \label{fig:halfcheetah}
\end{subfigure}\hspace*{\fill}
\begin{subfigure}{0.5\textwidth}
\includegraphics[width=1\columnwidth]{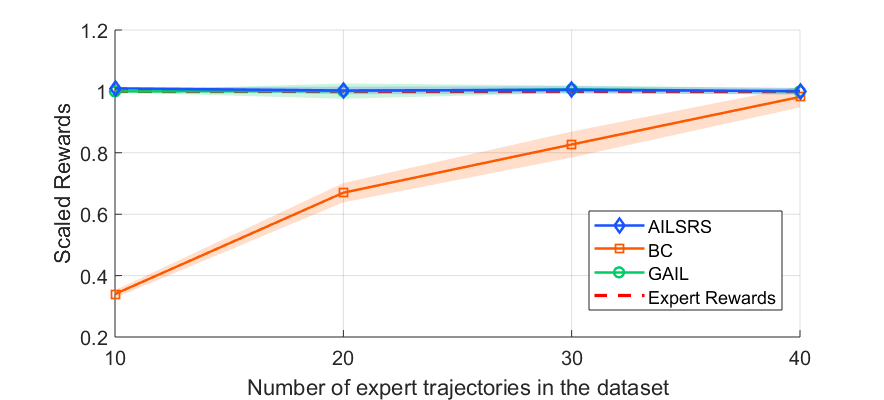}
\caption{Hopper-v2} \label{fig:hopper}
\end{subfigure}

\medskip
\begin{subfigure}{0.5\textwidth}
\includegraphics[width=1\columnwidth]{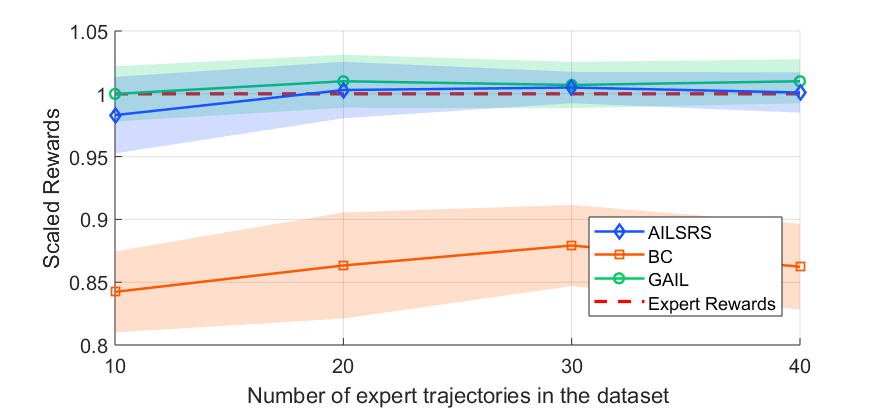}
\caption{Walker-v2} \label{fig:walker}
\end{subfigure}\hspace*{\fill}
\begin{subfigure}{0.5\textwidth}
\includegraphics[width=1\columnwidth]{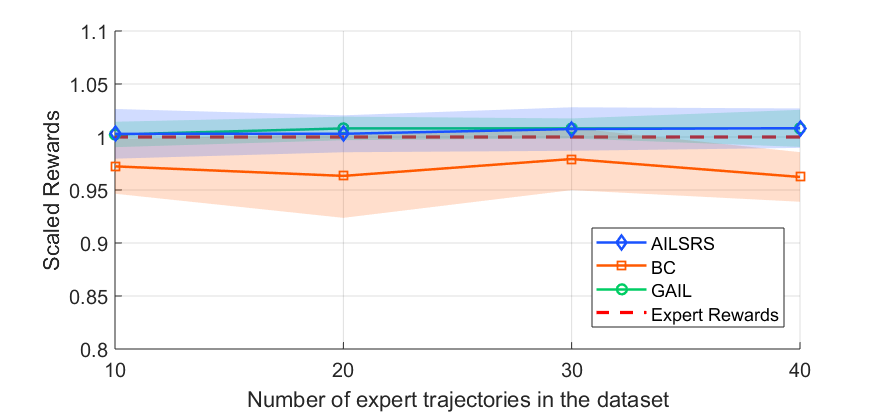}
\caption{Swimmer-v2} \label{fig:swimmer}
\end{subfigure}

\caption{The performance of  trained policy according to the set number of expert trajectories} \label{fig:2}
\end{figure*}
% Sample number based result ===================================================================

\BfPara{Sample Efficiency Experiments}
The purpose of experiments of Fig. \ref{fig:2} was to show the sample efficiency of AILSRS. In Fig. \ref{fig:2}, the blue lines means that the performance of the trained policies by AILSRS for the MuJoCo locomotion tasks. To compare the difference between the performance of AILSRS and GAIL, the experiments is executed on the HalfCheetah-v2, Swimmer-v2, Hopper-v2, and Walker-v2. In this experiment, in order to assess performance variability, repeated-measures were used based on three random seeds. Behavior cloning (BC) is used to accelerate the training policies for AILSRS and GAIL on HalfCheetah-v2 experiments. 
We evaluate AILSRS against two benchmars:
\begin{itemize}
    \item Behavior Cloning (BC) :  The policy is trained with supervised learning, using Adam optimizer. The policy parameter is trained to satisfy Equation (\ref{bc_objective}). 
    \item Generative Adversarial Imitation Learning (GAIL) : The algorithm of \cite{ho2016generative} using the objective function (\ref{gail_loss}). The implementation is based on OpenAI baseline with deterministic/stochastic policy GAIL~\cite{baselines}.
\end{itemize}
Fig. (\ref{fig:2}) presents that AILSRS achieve reasonable performance on MuJoCO locomotion tasks. On average, the performance of AILSRS were shown to learn stable policies. 

% HalfCheetah-v2 실험에서 평균 reward가 4632인 expert trajectories를 사용하였습니다. 이때, BC는 expert trajectories가 가장 적은 10개인 경우 1000 +- 10.32의 좋지 못한 성능을 보입니다. 그러나 expert trajectories의 갯수가 증가할 수록 뛰어난 성능을 보이며, 최종적으로는 4120 +- 129.12의 expert rewards에 근사한 성능을 보여줍니다. 그러나, 성능의 표준 편차가 매우 크며 expert rewards에는 도달하지 못하는 모습을 보여줍니다. 
% HalfCheetah-v2의 경우 GAIL과 AILSRS 모두 랜덤하게 초기화 된 정책을 기반으로 학습을 진행하면 실험환경에서 쉽게 좋은 결과를 얻어낼 수 없었습니다. 그래서, BC를 통해서 policy를 일정 부분 학습 한 후에 GAIL을 사용하여 imitation learning을 진행하였습니다. 그 결과, GAIL은 실험 환경에서 가장 적은 10개의 expert trajectories 에서도 expert rewards에 도달하는 성능을 보여줍니다. 그리고 expert trajectories가 증가함에 따라서 꾸준히 expert rewards보다 좋은 성능을 보여줍니다. 그리고 BC와 다르게 성능의 표준 편차도 매우 작은 모습을 보여주며 안정적인 정책이 학습되었음을 알 수 있습니다. 
% AILSRS도 GAIL과 마찬가지로 BC를 통해서 policy를 사전 학습 한 후에 AILSRS를 사용하여 imitation learning을 하였습니다. HalfCheetah-v2의 환경에서는 AILSRS도 역시 적은 expert trajectories에서도 expert rewards인 4632에 도달하였고 적은 표준 편차의 성능을 보여줍니다. 뿐만 아니라, BC와 달리 환경과 상호작용하며 학습하는 것으로 인해 더 뛰어난 성능을 보여주는 것을 확인할 수 있습니다. 

In the HalfCheetah-v2 experiment, we used expert trajectories with an average reward of 4632. At this time, BC has the worst performances of 1000 $\pm$ 10.32 when 10 expert trajectories are the least. However, as the number of expert trajectories increases, performance is better, and finally, performance is better than 4120 $\pm$ 129.12 expert rewards. However, the standard deviation of the performance is very large and shows that it does not reach the expert rewards.

In the case of HalfCheetah-v2, I used imitation learning using GAIL after learning some policy through BC. As a result, GAIL shows the ability to reach expert rewards in the lowest 10 expert trajectories in the experimental environment. And as expert trajectories increase, they perform consistently better than expert rewards. And, unlike BC, the standard deviations of performance are very small, and we can see that stable policy has been learned.

As with GAIL, AILSRS pre-learned the policy through BC and imitation learning was done using AILSRS. In the HalfCheetah-v2 environment, AILSRS also reached 4632, the expert rewards in lesser expert trajectories, and shows less standard deviation performance. In addition, unlike BC, you can see that it performs better by interacting with the environment and learning.

% ============================= 아래는 아직 분량을 정하기 위해 아무 영어나 넣은 내용입니다 =========
% Hopper

% Hopper-v2 실험에서 평균 reward가 3245인 expert trajectories를 사용하였습니다. 이때, BC는 expert trajectories가 가장 적은 10개인 경우 1200 +- 10.325의 좋지 못한 성능을 보입니다. Expert trajectories의 갯수가 증가할 수록 HalfCheetah-v2에서와 마찬가지로 최종적으로는 3224 +- 15.413의 expert rewards에 준하는 성능을 보여줍니다. Hopper-v2같은 경우 HalfCheetah-v2에 비해서 BC도 성능면에서 안정적인 모습을 보여줍니다. 
% Hopper-v2의 실험에서 GAIL과 AILSRS 모두 BC를 사용하지 않고 학습을 진행하였습니다. GAIL은 가장 적은 expert trajectories 10개 부터 안정적으로 expert rewards에 준하는 성능을 보여줍니다. 
% AILSRS도 GAIL과 마찬가지로 적은 expert trajectories에서도 expert rewards에 도달하는 안정적인 성능을 보여줍니다. 

We used expert trajectories with an average reward of 3245 in the Hopper-v2 experiment. At this time, BC has the worst performances of 1200 $\pm$ 10.325 when 10 expert trajectories are the least. As the number of Expert trajectories increases, the performance is similar to the expert rewards of 3224 $\pm$ 15.413, as in HalfCheetah-v2. In the case of Hopper-v2, BC is more stable than HalfCheetah-v2.
In the Hopper-v2 experiment, both GAIL and AILSRS did not use BC. GAIL has the lowest number of expert trajectories (10), and it shows stable performance comparable to expert rewards.
AILSRS, like GAIL, shows stable performance reaching expert rewards in fewer expert trajectories.

% Walker & Swimmer

% Walker의 경우에는 1021의 expected rewards를 보여주는 expert trajectories를 사용하였으며, Swimmer의 경우 362의 expected rewards를 보여주는 데이터를 사용하였습니다. 
% Walker와 Swimmer는 HalfCheetah-v2와 Hopper-v2에 비해서 3가지 benchmark 알고리즘 모두가 좋은 성과를 보여주는 것을 알 수 있습니다. Walker-v2의 실험에서 BC는 10개의 expert rewards일 때는 862.745 +-  30.62 정도의 성능을 보여주며, 평균적으로 가장 좋은 성능을 보인 30개의 expert trajectories를 기반으로 학습하였을 때 890.823 +- 31.23 의 성능을 보여주고 있습니다. 모든 실험에서  다른 알고리즘들 보다 다소 성능이 떨어지며며 학습된 정책이 보여주는 성능의 안정성 또한 떨어지는 모습을 보여줍니다. GAIL과 AILSRS의 경우 모두 Expert rewards에 1021 +- 51.05 정도의 성능 범위 안에서 효과적으로 정책을 학습하는 결과를 보여주고 있습니다.
Expert trajectories were used to show expected rewards of 1021 for Walker, and data showing 362 expected rewards for Swimmer. Walker and Swimmer show that all three benchmark algorithms perform well compared to HalfCheetah-v2 and Hopper-v2. In the Walker-v2 experiment, BC showed performance of 862.745 $\pm$ 30.62 for 10 expert rewards, and 890.823 $\pm$ 31.23 for 30 expert trajectories on average. I give. In all experiments, performance is somewhat less than the other algorithms, and the performance of the learned policy is also less stable. In both GAIL and AILSRS, we show that we are effectively learning policy within the performance range of 1021 $\pm$ 51.05 for Expert rewards.

% Swimmer-v2의 실험에서는 Walker-v2처럼 3가지 benchmark 알고리즘 모두가 좋은 성과를 보여줍니다. BC는 10개의 expert trjaectories에서 352.95 +- 9.16의 성능을 보여주며 이전의 실험들 보다 가장 expert의 성능에 근접한 결과를 보여줍니다. 나머지 Swimmer-v2의 실험에서도 유사한 좋은 성능을 보여주고 있습니다.
In the Swimmer-v2 experiment, all three benchmark algorithms, such as Walker-v2, show good performance. BC shows performance of 352.95 + - 9.16 in 10 expert trjaectories, and is closer to the performance of the most expert than previous experiments. The rest of the Swimmer-v2 experiment shows similar good performance.

% GAIL 과 AILSRS 다른 실험과 마찬가지로 expert rewards에 준하는 성능을 보여주고 있습니다. 그와 함꼐 4.51 정도의 작은 표준편차를 보이며 안정적인 학습 성은을 보여주고 있습니다. 

GAIL and AILSRS Like other experiments, it shows performance equivalent to expert rewards. At the same time, it shows a small standard deviation of about 4.51 and shows stable learning performance.

% 이 Section을 통해서 우리는 BC와 GAIL과의 비교를 통해서 제안한 AILSRS의 performance를 보였습니다. 그 결과 AILSRS는 parameter space of the policies의 방법으로 highdimensional Mujoco locomotion tasks에서 environment의 직접적인 reward 없이 expert trjectories를 기반으로 simple linear policy를 성공적으로 학습하는 모습을 보여주었다. 
Overall, these results indicate that the random search in the parameter space of policies can be used to imitation learning. Through this section we demonstrate that the performance of the proposed AILSRS shows competitive performance comparing with BC and GAIL. AILSRS showed the successful learning of expert trajectories without direct reward of environment on MuJoCo locomotion tasks. Together these results provide important possibility into the imitation learning using random search with simple linear policies.

% ===========================================================================================
\begin{figure*}[t] 
\begin{subfigure}{0.5\textwidth}
\includegraphics[width=1\columnwidth]{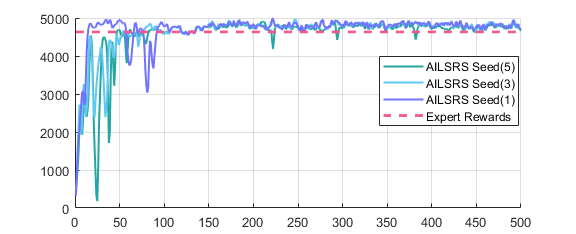}
\caption{HalfCheetah-v2} \label{fig:halfcheetah}
\end{subfigure}\hspace*{\fill}
\begin{subfigure}{0.5\textwidth}
\includegraphics[width=1\columnwidth]{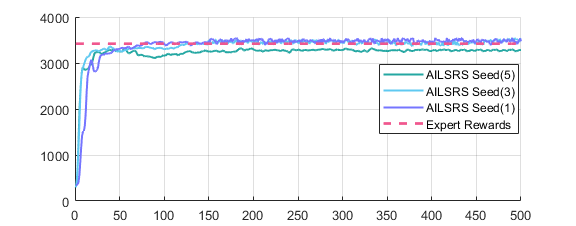}
\caption{Hopper-v2} \label{fig:hopper}
\end{subfigure}

\medskip
\begin{subfigure}{0.5\textwidth}
\includegraphics[width=1\columnwidth]{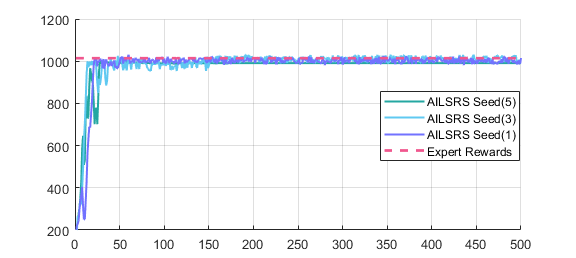}
\caption{Walker-v2} \label{fig:walker}
\end{subfigure}\hspace*{\fill}
\begin{subfigure}{0.5\textwidth}
\includegraphics[width=1\columnwidth]{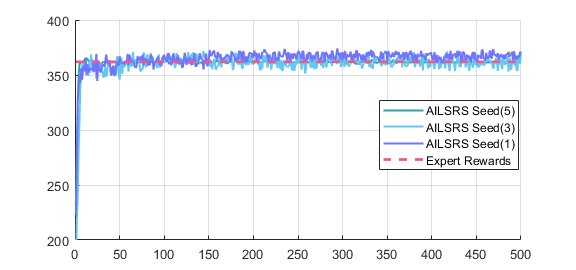}
\caption{Swimmer-v2} \label{fig:swimmer}
\end{subfigure}

\caption{An evaluation of AILSRS on the OpenAI Gym and mujoco locomotion tasks. The training curves are averaged for each random seed experiments.} 
\label{fig:3}
\end{figure*}
% ===========================================================================================

\BfPara{Training Curve}
The this sub section of the experiments was concerned with the training stability when we use multiple random seeds.
% Fig. \ref{fig:3}은 AILSRS의 Mujoco locomotion tasks의 training curve를 보여줍니다. 각 실험은 random seed의 숫자를 증가시켜가면서 학습할 때의 결과를 보여주고 있으며, 20번의 실험 결과들 중 outlier들을 제외한 나머지 실험들 중 랜덤하게 선택한 5번의 실험 결과의 평균을 Gaussian filter를 통해서 smooth 하게 만든 그래프 입니다.
Fig. \ref{fig:3} shows the training curve of Mujoco locomotion tasks in AILSRS. Each experiment shows the results of learning by increasing the number of random seeds. The average of 5 randomly selected experimental out of 20 experimental results is smoothed through a Gaussian filter. Graph made.

% HalfCheetah-v2의 경우, random seed가 1개, 3개 그리고 5개 일때 모두 expert rewards에 도달하는 모습을 보여준다. 그러나, 학습 초기 random seed의 수가 증가할 수록 학습 과정에서 불안정함을 보여준다. Hopper-v2의 경우, random seed가 1개 그리고 3개 일때 expert rewards에 도달하는 모습을 보여주지만, 5개 일때는 조금 못 미치는 결과를 보여준다. Walker-v2와 Swimmer-v2의 경우, 모든 실험 결과가 expert rewards에 도달하는 모습을 보여준다. 이 그래프를 통해서 AILSRS가 random seed가 1개, 3개 그리고 5개로 증가하여도 빠르게 성공적으로 expert trajectories로 부터 환경의 직접적인 reward signal 없이 정책을 학습할 수 있다는 것을 보여줍니다. 그러나, Mujoco locomotion tasks에서 어려운 문제인 Ant-v2와 Humanoid-v2의 경우 학습 속도가 굉장히 더딘 모습을 보였으며, 해당 환경에 대해서는 아직 환경과 제안된 알고리즘에 대한 분석이 필요하다.
In the case of HalfCheetah-v2, it shows that the trained policies in 1, 3 and 5 random seed environments reaches the expert's reward. However, as the number of random seeds increases, the learning process becomes unstable. In case of hopper-v2, it shows that reaching expert rewards when 1 and 3 random seeds are reached, but it is slightly less than 5 when it is 5. For Walker-v2 and Swimmer-v2, all experimental results show that they reach expert rewards. This graph shows that AILSRS can quickly and successfully learn policies from expert trajectories without a direct reward signal of the environment, even if the number of random seeds increases to one, three, and five. However, in the case of Ant-v2 and Humanoid-v2 which are difficult problems in Mujoco locomotion tasks, the learning speed is very slow, and the environment and analysis of the proposed algorithm are still necessary for the environment.
\section{Concluding Remarks}\label{sec:final}
The proposed simple random search based imitation learning method is not only a derivative free but aloso model-free reinforcement learning algorithm. Furthermore, the simple linear policies are used to control complex dynamic tasks. It shows competitive performance on MuJoCo locomotion tasks. The simple update step makes the algorithm to be facile; and thus it makes the reconstruction results is able to get reasonable performance easily. 

By comparing the performance of the proposed model with complex deep reinforcement learning based imitation learning, we demonstrated that simple random search based imitation learning could be used to train linear policies that achieve reasonable performance on the MuJoCo locomotion tasks. 

This results can be a breakthrough to the common belief that random searches in the parameter space of policy can not be competitive in terms of performance. However, the proposed method was not able to get competitive performance on the Ant-v2 and Humanoid-v2 within reasonable training time. Therefore, since the proposed method is simple on-policy algorithm, we can perform extensive research as future research directions.

\section*{Acknowledgment}
This research was supported by the National Research Foundation of Korea (2016R1C1B1015406, 2017R1A4A1015675); and also by Institute for Information \& Communications Technology Promotion (IITP) grant funded by the Korea government (MSIT) (No.2018-0-00170, Virtual Presence in Moving Objects through 5G).
J. Kim is a corresponding author of this paper.

\bibliographystyle{IEEEtran}  
\bibliography{reference}

\vspace{12pt}

\end{document}